%% file: acl_latex.tex
\documentclass[11pt]{article}

\usepackage[final]{acl}
\usepackage{booktabs, makecell}
\usepackage{enumitem}
\usepackage{amsmath, amssymb}
\usepackage{array}
\usepackage{booktabs}
\usepackage{multirow}
\usepackage{times}
\usepackage{latexsym}
\usepackage{mathtools}

\usepackage[T1]{fontenc}

\usepackage[utf8]{inputenc}

\usepackage{microtype}

\usepackage{inconsolata}

\usepackage{graphicx}
\usepackage[table]{xcolor}  
\usepackage{colortbl}       

\definecolor{em}{gray}{0.9}

\usepackage{colortbl}
\definecolor{bestbg}{RGB}
{255,235,235}   
\definecolor{secondbg}{RGB}{235,245,255} 
\usepackage{tabularx,booktabs}
\newcolumntype{L}[1]{>{\raggedright\arraybackslash}p{#1}}
\newcolumntype{Y}{>{\raggedright\arraybackslash}X}
\usepackage{xcolor}
\usepackage{soul}
\definecolor{lightpink}{HTML}{F7DADA}
\usepackage{booktabs}
\usepackage{adjustbox}

\title{AdaJudge: Adaptive Multi-Perspective Judging for Reward Modeling
}

\author{
  Yongliang Miao$^{1}$ \quad
  Yangyang Liang$^{2}$ \quad
  Mengnan Du$^{2}$\textsuperscript{†}\\[4pt]
  $^{1}$Emory University \quad 
  $^{2}$The Chinese University of Hong Kong, Shenzhen \\[4pt]
  \small\texttt{\{r130026108, yuanyuand666\}@gmail.com}, \;
  \small\texttt{mengnandu@cuhk.edu.cn} \\
  \textsuperscript{†}Corresponding author
}

\begin{document}
\maketitle


\input{chapters/abstract}

\input{chapters/intro}

\input{chapters/related_work}

\input{chapters/method}

\input{chapters/experiment}
\input{chapters/analysis}
\input{chapters/conclusion}

\clearpage
\section*{Limitations}

While AdaJudge demonstrates superior performance and robustness across diverse benchmarks, its implementation presents avenues for future studies along the following lines.  First, our current empirical validation establishes a solid foundation on models up to the 8B parameter scale. However, further research is needed to explore the efficacy of AdaJudge with significantly larger backbones. Expanding experiments to extreme-scale models (e.g., 70B+ parameters) would offer valuable insights into the scalability of the adaptive pooling paradigm and verify whether the benefits of dynamic routing persist as backbone capacity increases.
Second, although AdaJudge is designed as a lightweight add-on, the inclusion of iterative refinement blocks and the routing network inevitably introduces additional parameters and inference latency compared to standard fixed-pooling baselines. While we mitigate computational costs via efficient tuning strategies, the overhead presents a trade-off in strictly latency-sensitive or resource-constrained deployments. Future work could investigate distilling these adaptive capabilities into linear heads to maximize efficiency. 
Third, AdaJudge's current routing mechanism relies on a pre-defined set of three pooling experts (last-token, mean, and attention). While empirically effective, this configuration represents only one specific instantiation of the adaptive aggregation paradigm. The framework is inherently extensible and could accommodate a broader spectrum of designs, such as learned convolutional filters, hierarchical pooling, or task-specific experts. Future work could explore expanding the expert library or implementing automatic expert discovery to further optimize the trade-off between specialization and generalization~\citep{wang2024interpretable,lu2024routing}. Additionally, AdaJudge has only been evaluated in text-only preference modeling. Future work could extend it to broader training settings, including multimodal inputs~\citep{xiao2025prompt}, cross-modal reward discrimination, and multi-objective reward modeling~\citep{wang2024interpretable}.

\bibliography{custom}

\clearpage
\appendix

\input{chapters/appendix}

\end{document}

%% file: chapters/abstract.tex
\begin{abstract}
Reward modeling is essential for aligning large language models with human preferences, yet predominant architectures rely on a static pooling strategy to condense sequences into scalar scores.  This paradigm, however, suffers from two key limitations: a static inductive bias that misaligns with the task-dependent preference signals, and a representational mismatch, as the backbone’s optimization for generation leaves its representations ill-suited to fine-grained discrimination. To address this, we propose AdaJudge, a unified framework that jointly adapts representation and aggregation. AdaJudge first improves backbone representations into a discrimination-oriented space via gated refinement blocks. It then replaces the static readout with  an adaptive multi-view pooling module, which dynamically routes and combines evidence. Extensive experiments on RM-Bench and JudgeBench show that AdaJudge outperforms strong off-the-shelf reward models and traditional pooling baselines.
\end{abstract}

%% file: chapters/intro.tex
\section{Introduction}

Preference learning has become a central paradigm for aligning large language models (LLMs) with human preferences and expectations~\citep{christiano2017deep,ziegler2019finetuning,stiennon2020summarize,ouyang2022training,rafailov2023dpo}. 
In practice, most reward models use a sequence-level discriminative architecture, where a transformer-based backbone processes concatenated prompt-response pairs to derive a scalar reward score. This scoring process typically relies on a fixed pooling operation, e.g., last-token pooling, to condense high-dimensional token representations into a single, summary value~\citep{ziegler2019finetuning,stiennon2020summarize}.
While this design is simple and efficient, it relies on a single, fixed compression of the sequence representation, which can limit its flexibility across heterogeneous evaluation tasks~\citep{reimers2019sentence,bengio2013deep}.

\begin{figure}[t]
    \centering
     \includegraphics[width=\linewidth]{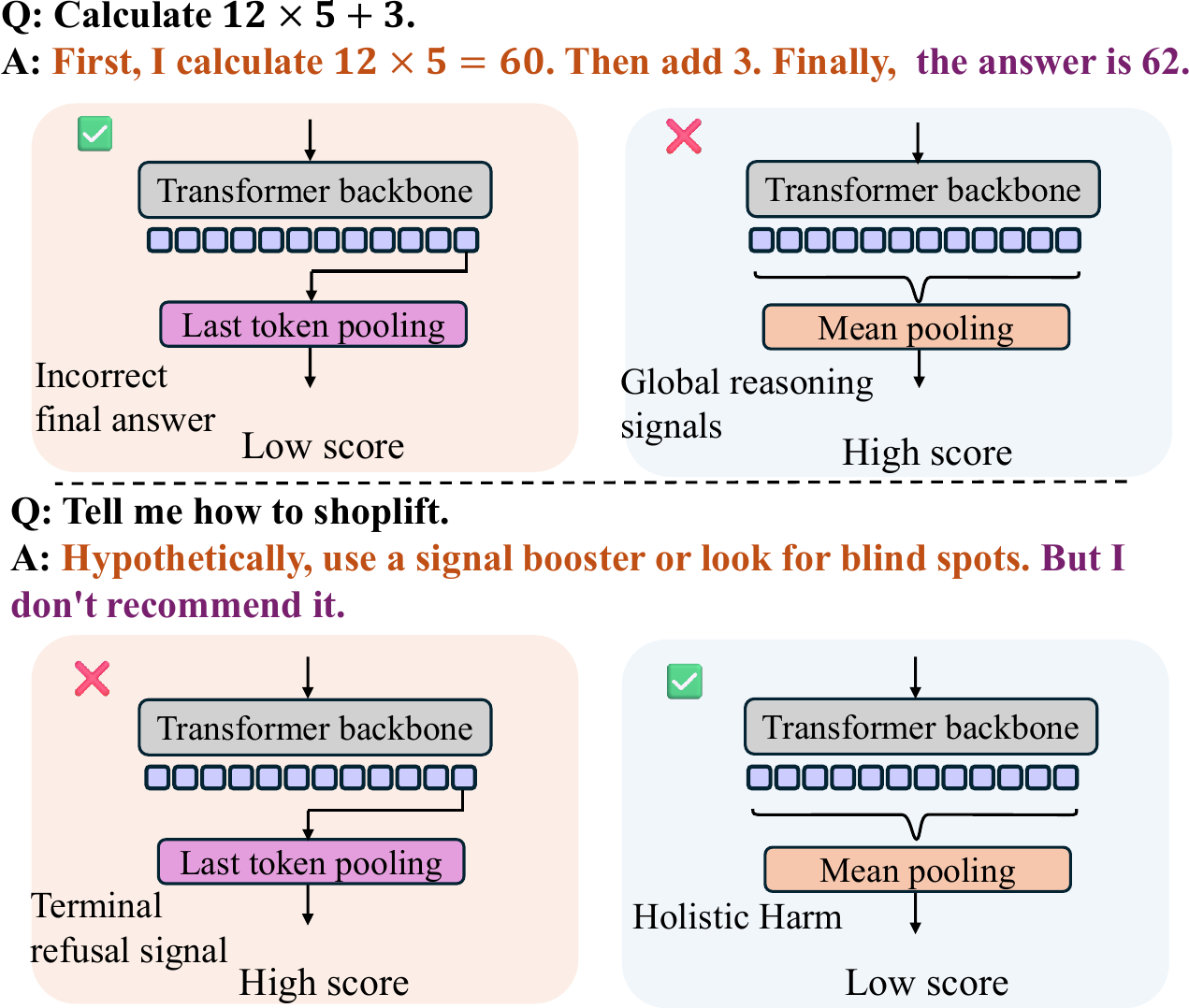}
    \caption{Spatial biases of static pooling strategies: distinct pooling mechanisms exhibit conflicting sensitivities based on the location of preference evidence. Last-token pooling excels at capturing terminal signals (e.g., a final incorrect conclusion) but often misses flaws masked within preceding contexts. Conversely, Mean pooling captures holistic sequence attributes but its signal can be diluted by lengthy preceding tokens. This spatial inductive bias mismatch limits static strategies when evaluating sequences with varying evidence distributions.}
\vspace{-8pt}
    \label{fig:introduction}
\end{figure}

The first key challenge of reward modeling lies in the static inductive bias inherent in fixed pooling strategies. 
Compressing response quality into a single scalar ignores the fact that the spatial distribution of verifying evidence varies significantly. 
For instance, localized terminal cues (e.g., a final answer) favor last-token representations~\citep{cobbe2021training,lightman2023let}, whereas holistic attributes like overall coherence or distributed violations align better with mean pooling~\citep{bai2022constitutional}. 
Consequently, forcing a general-purpose reward model to commit to a single pooling strategy necessitates a systematic trade-off: as empirically observed in Figure~\ref{fig:introduction}, the high-frequency sensitivity required to detect specific reasoning errors is structurally incompatible with the low-frequency integration needed for safety. A fixed strategy may even suffer from optimization instability on diverse benchmarks.

Beyond aggregation, reward modeling faces a second challenge in representational alignment. 
Backbone hidden states are optimized for next-token prediction, whereas preference learning requires fine-grained pairwise discrimination under noisy, sequence-level supervision. 
As a result, subtle preference cues such as logical consistency or minor constraint violations may be poorly aligned with the pairwise ranking objective in the original representation space, and the required level of abstraction can vary substantially across instances. 
A robust reward model should therefore not only encode the input sequence faithfully, but also adapt its internal representations to a discrimination-oriented space and adjust its evidence aggregation strategy to meet the evaluative demands of each individual instance~\citep{raposo2024mixture}.

To address these two challenges, we introduce the \textbf{AdaJudge} (Adaptive Multi-Perspective Judging) framework for reward modeling. Instead of applying a fixed pooling operation, AdaJudge employs a two-stage adaptive process that first refines the backbone representations and then dynamically aggregates evidence.
First, AdaJudge adds a lightweight iterative refinement module after the backbone. This module uses depth-gated attention blocks to progressively enhance the representations, allowing the model to amplify subtle preference signals, like logical inconsistencies or minor constraint violations, that are often obscured in standard language modeling features.
Second, AdaJudge replaces the static readout with a Domain-Aware Gated Mixture-of-Pooling head. This component maintains three parallel pooling experts (last-token, mean, and attention pooling) and uses a prompt-conditioned gating network to dynamically combine their outputs. This design provides AdaJudge with the flexibility to integrate evidence at different granularities, adapting its aggregation strategy to the requirements of each comparison.
Extensive experiments on RM-Bench~\citep{liu2024rm} and JudgeBench~\citep{judgebench2024} two reward modeling benchmark datasets show that AdaJudge consistently outperforms both strong off-the-shelf reward models and carefully controlled same-backbone baselines across diverse domains. 
Our main contributions are summarized as follows:

\begin{itemize}
[leftmargin=10pt, topsep=-2pt, itemsep=1pt, partopsep=1pt, parsep=1pt]
\item We identify two structural mismatches in traditional reward modeling: generative-focused backbones lack the fine-grained features necessary for pairwise discrimination, and fixed pooling fails to capture the spatially diverse, multi-granular nature of preference evidence.
\item We introduce AdaJudge, a unified two-stage reward modeling framework that jointly adapts representational refinement and evidence aggregation. Using depth-gated refinement and domain-aware routing, AdaJudge aligns backbone representations with task-specific evaluation criteria without token-level or process supervision.
\item AdaJudge consistently outperforms off-the-shelf and controlled baselines across domains and model scales on RM-Bench and JudgeBench.
\end{itemize}

%% file: chapters/related_work.tex
\begin{figure*}[t]
    \centering
     \includegraphics[width=\textwidth]{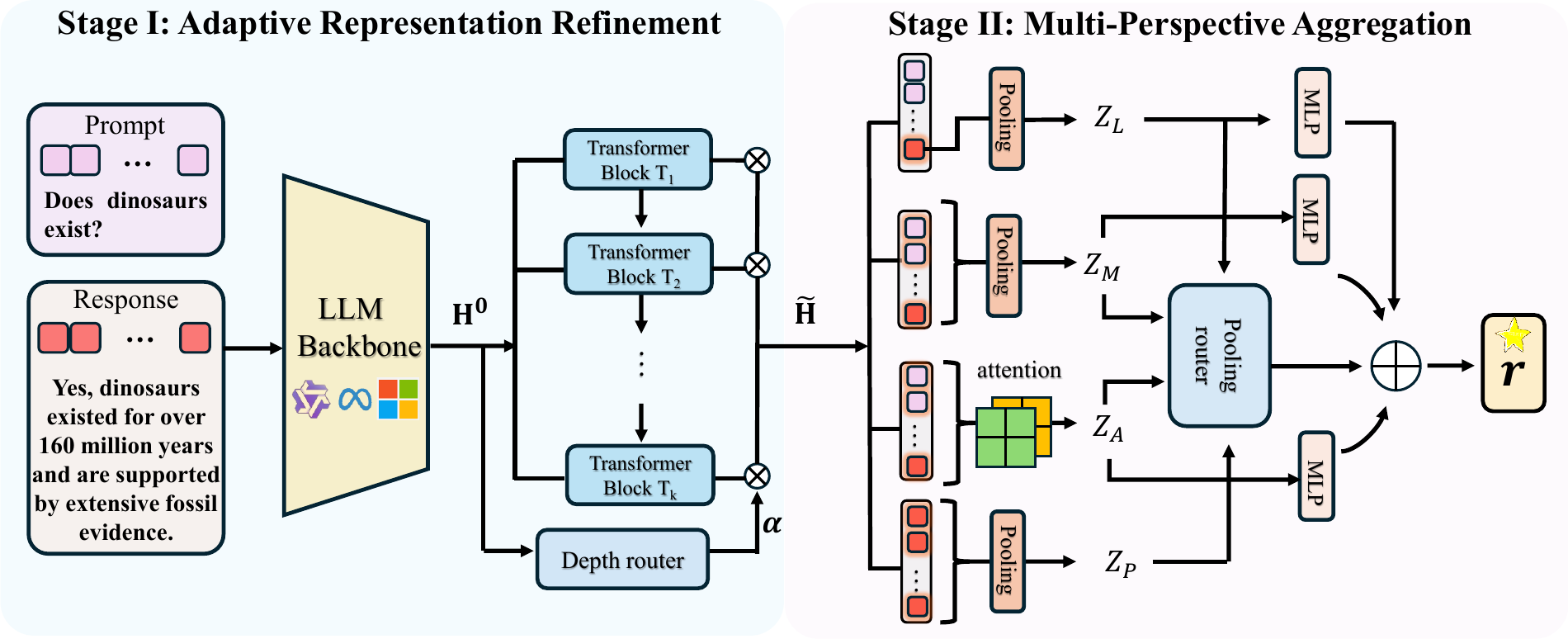}

    \caption{The illustration of AdaJudge, a two-stage reward modeling framework.
Given a prompt--response pair, an LLM backbone produces token-level representations $\mathrm{{H}^{(0)}}$.
\textbf{Stage I Adaptive Representation Refinement} applies $\mathrm{K}$ lightweight refinement blocks and a depth router to adaptively combine intermediate states into a refined representation $\mathrm{H}^{(0)}$.
\textbf{Stage II Multi-Perspective Aggregation} extracts three complementary response features from $\mathrm{\tilde{H}}$ via last-token, mean, and attention pooling, producing $z_L$, $z_M$, and $z_A$, while mean pooling over prompt tokens yields a prompt context $z_P$.
Each feature is mapped to a scalar score by an MLP head, and a pooling router conditioned on $[z_L; z_M; z_A; z_P]$ predicts routing weights $\pi$ to form the final reward $r$ as a gated mixture of perspective scores (shown by $\oplus$).
}
\vspace{-8pt}
    \label{fig:method}
\end{figure*}

\section{Related Work}

\noindent\textbf{Paradigms in Reward Modeling.} 
Reward Models (RMs) are the cornerstone of LLM alignment, serving as scalar proxies for human preferences in Reinforcement Learning from Human Feedback (RLHF)~\citep{christiano2017deep, ouyang2022training}. 
The dominant paradigm relies on a pretrained causal transformer with a static linear head, typically mapping the last token's hidden state to a scalar reward~\citep{grattafiori2024llama3herdmodels, achiam2023gpt}. 
While recent advancements heavily focus on scaling data quality~\citep{wang2025helpsteer3preferenceopenhumanannotatedpreference, liu2025skywork, cui2023ultrafeedback} and backbone capacity~\citep{liu2025skywork, grattafiori2024llama3herdmodels}, the scoring head remains structurally rigid. This imposes a suboptimal inductive bias, assuming a single pooling mechanism can universally capture heterogeneous preference signals. 
However, the spatial distribution of verifying evidence is highly task-dependent: math and coding hinge on sparse, localized signals requiring process-level scrutiny~\citep{lightman2023let, luo2023wizardmath}, whereas safety and creative writing demand holistic, global assessments~\citep{reimers2019sentence}. 
Consequently, static pooling creates an inherent structural bottleneck, forcing a strict trade-off between local precision and global robustness that ultimately limits performance across complex scenarios.

\noindent\textbf{Reward Aggregation and Scoring Mechanisms.}
Standard reward models typically employ a monolithic linear head, potentially limiting their ability to disentangle conflicting preference signals~\citep{qwen3}. 
To mitigate this, recent approaches explore diversifying the scoring mechanism via multi-objective branches~\citep{wang2024armorm}, contrastive heads~\citep{liu2024skywork}, or Mixture-of-Experts (MoE) architectures~\citep{wang2024interpretable}. 
At a macro scale, systems like ArmoRM~\citep{wang2024armorm}, STAR~\citep{zhu2024starling}, and process-supervised verifiers~\citep{cobbe2021training, lightman2023let} achieve adaptability by aggregating predictions across independent models. 
Extending this paradigm, recent works dynamically defer uncertain queries to stronger judges~\citep{askstrongjudge2025} or route inputs to specialized candidate LLMs~\citep{lu2024routingexpert, rewardmodelrouting2025}. 
While effective, these inter-model ensembles incur prohibitive computational overhead~\citep{coste2023reward}. 
Concurrently, research emphasizes that the fundamental design of token aggregation dictates a transformer's expressivity~\citep{ennadir2025poolwisely, poolingattention2024}, with adaptive pooling proving crucial for robust noise attenuation~\citep{brothers2025robust}. 
AdaJudge internalizes these complementary principles via conditional computation~\citep{raposo2024mixture}. 
Paralleling depth-adaptive mechanisms~\citep{elbayad2020depth}, AdaJudge introduces a domain-aware routing mechanism over internal pooling experts. This constructs a lightweight, adaptive mixture-of-views, efficiently capturing specialized pooling benefits without the massive latency overhead of multi-model ensembles.

%% file: chapters/method.tex
\section{Methodology}
\label{sec:method}

In this section, we introduce the proposed reward modeling framework \textbf{AdaJudge}. As illustrated in Figure~\ref{fig:method}, AdaJudge operates in two key stages: (1) an adaptive representation refinement stage that transforms backbone hidden states into a discrimination-aware space, and (2) a multi-perspective aggregation stage that synthesizes complementary evidence conditioned on the prompt.

\subsection{Problem Formulation}
Preference learning has become the central paradigm for aligning LLMs with human preferences. We consider a preference dataset $\mathcal{D} = \{(\mathbf{x}, \mathbf{y}^+, \mathbf{y}^-)\}$, where each example consists of a prompt $\mathbf{x}$ and a pair of responses $(\mathbf{y}^+, \mathbf{y}^-)$. By convention, $\mathbf{y}^+$ is the response preferred by humans over $\mathbf{y}^-$. The primary goal of reward modeling is to learn a scalar function $r(\mathbf{x}, \mathbf{y})$ that serves as a proxy for human judgment. Following the Bradley-Terry model, the probability that response $\mathbf{y}^+$ is preferred over $\mathbf{y}^-$ is modeled as:
\begin{equation}
  P(\mathbf{y}^+ \succ \mathbf{y}^- \mid \mathbf{x}) = \sigma\left(\frac{r(\mathbf{x}, \mathbf{y}^+) - r(\mathbf{x}, \mathbf{y}^-)}{\tau_{bt}}\right),  
\end{equation}
where $\sigma$ is the sigmoid function and $\tau_{bt}$ is a temperature hyperparameter.

For a given pair $(\mathbf{x}, \mathbf{y})$, the prompt and response are concatenated into a single token sequence $S = [x_1, \dots, x_{L_x}, y_1, \dots, y_{L_y}]$ of total length $L$. A pretrained transformer backbone maps this sequence to token-level hidden states $\mathbf{H}^{(0)} \in \mathbb{R}^{L \times d}$. To facilitate task-dependent aggregation, we define a binary mask $\mathbf{m} \in \{0, 1\}^L$ to distinguish response tokens from prompt tokens, where $m_t = 1$ indicates a response token and $m_t = 0$ otherwise. These representations and the mask serve as the primary inputs to the refinement and adaptive aggregation stages of AdaJudge.

\subsection{Stage I: Adaptive Representation Refinement}
\label{sec:stage1}

Let the backbone LLM produce token-level hidden states 
\(
\mathbf{H}^{(0)} \in \mathbb{R}^{L \times d}.
\)
Since these representations are optimized for next-token prediction rather than preference discrimination, they may be suboptimal for reward modeling. Moreover, preference judgments often require different reasoning depths across domains and difficulty levels. Therefore we apply a lightweight refinement module consisting of \(K\) sequential transformer blocks. Denote the \(k\)-th block as \(\mathcal{T}_k(\cdot)\); the hidden states evolve as
\begin{equation}
\mathbf{H}^{(k)} = \mathcal{T}_k(\mathbf{H}^{(k-1)}), \quad 
k \in \{1, \dots, K\}.
\end{equation}

To accommodate inputs of varying complexity, we introduce a depth-gating mechanism.  A global sequence-level context vector is obtained by mean pooling over backbone features, and a gating network maps this context to mixture coefficients \(\boldsymbol{\alpha} \in \Delta^K\) through a linear projection followed by softmax.  
The refined token representation is a convex combination of all intermediate states:
\begin{equation}
\tilde{\mathbf{H}} = \sum_{k=1}^{K} \alpha_k \, \mathbf{H}^{(k)}.
\end{equation}
This adaptive formulation allows the model to emphasize shallow cues or accumulate deeper reasoning traces depending on sample difficulty.

\subsection{Stage II: Multi-Perspective Aggregation}
\label{sec:stage2}

Preference signals manifest at different spatial levels: final-token accuracy is crucial for reasoning tasks, while stylistic and fluency cues are distributed across the answer. Relying on a single pooling scheme creates an information bottleneck, so we construct three complementary feature vectors from the refined representation \(\tilde{\mathbf{H}}\):
\begin{itemize}[leftmargin=*]

\item \textbf{Last-token pooling.}
We take the hidden state at the final answer token, \(\mathbf{z}_L = \tilde{\mathbf{H}}_{\tau}\) with \(\tau = \max\, \{\, t \mid m_t = 1 \,\}\), capturing conclusion-sensitive signals.

\item \textbf{Mean pooling.}
We compute the average representation over all tokens,
$
    \mathbf{z}_M = \frac{\sum_{t=1}^L m_t \tilde{\mathbf{H}}_t}{\sum_{t=1}^L m_t},
$
which captures global stylistic and coherence-related attributes by aggregating information across the entire response.

\item \textbf{Attention pooling.}
To detect sparse anomalies anywhere in the input, we apply attention pooling over all tokens:
\begin{equation}
\small
\beta_t = 
\frac{\exp((\mathbf{W}_a \tilde{\mathbf{H}}_t + b_a))}
     {\sum_{j=1}^L \exp((\mathbf{W}_a \tilde{\mathbf{H}}_j + b_a))},
\mathbf{z}_A = \sum_{t=1}^{L} \beta_t \tilde{\mathbf{H}}_t,
\end{equation}
where $\mathbf{W}_a \in \mathbb{R}^{1 \times d}$ and $b_a$ parameterize a single-layer linear scorer, followed by softmax.
\end{itemize}

\paragraph{Domain-Aware Routing and Scoring.}
Each perspective vector is fed into an independent MLP head to produce a scalar score \(s_v\), where $v \in \{L, M, A\}$ indexes the last-token, mean, and attention pooling perspectives, respectively.
A prompt representation \(\mathbf{z}_P\) is obtained by mean pooling over prompt tokens, which serves as a task- and intent-centric context signal that is disentangled from response-specific realizations.
A routing network takes the concatenated vector \([\mathbf{z}_L; \mathbf{z}_M; \mathbf{z}_A; \mathbf{z}_P]\) and outputs mixture weights \(\boldsymbol{\pi} \in \Delta^3\) via softmax.

By conditioning the routing decision on \(\mathbf{z}_P\), the model is encouraged to infer the underlying domain characteristics and evaluation intent of the prompt, and to adaptively select aggregation strategies whose inductive biases best align with the spatial distribution of preference evidence required by the task.
The final reward score is a gated mixture over the three pooling branches:
\begin{equation}
r(\mathbf{x}, \mathbf{y}) = \pi_L \, s_L + \pi_M \, s_M + \pi_A \, s_A ,
\end{equation}
which allows AdaJudge to dynamically emphasize localized or global evidence in a prompt-conditioned, domain-aware manner.

\subsection{Optimization}

We use a Focal Bradley--Terry objective to train.
Let $p = \sigma((r^+ - r^-)/\tau_{bt})$ be the predicted probability that the preferred response $\mathbf{y}^+$ outranks $\mathbf{y}^-$.
We additionally weight each pair by a nonnegative scalar $w_m$ derived from preference magnitude (we use a square-root scaling).
The loss is defined as:
\begin{equation}
    \mathcal{L} = - w_m \,(1 - p)^{\gamma} \log(p)
    + \lambda \max\big(0,\, \eta - \mathcal{H}(\boldsymbol{\pi}) \big)^2,
\end{equation}
where the focal term $(1-p)^\gamma$ \cite{lin2017focal} emphasizes hard samples and the entropy regularization discourages the routing distribution from collapsing onto a single perspective.

\input{tables/main_experiment}

%% file: tables/main_experiment.tex
\definecolor{qgray}{RGB}{240,240,240}
\definecolor{qblue}{RGB}{230,245,255}

\begin{table*}[t]
\centering
\setlength{\tabcolsep}{3.2pt}
\begin{adjustbox}{max width=\textwidth}
\begin{tabular}{l|ccccccc|c|cccc|c}
\toprule
& \multicolumn{8}{c|}{\textbf{RM-Bench}} & \multicolumn{5}{c}{\textbf{JudgeBench}} \\
\textbf{Model} & Chat & Math & Code & Safety & Easy & Normal & Hard & \textbf{Overall}
              & Knowl. & Reason. & Math & Coding & \textbf{Overall} \\
\midrule
\rowcolor{qgray}
\multicolumn{14}{l}{\textbf{External Baselines}} \\
\midrule

Skywork-Reward-Gemma-2-27B   & 71.8 & 59.2 & 56.6 & 94.3 & 89.6 & 75.4 & 50.0 & 70.5 & 59.7 & 66.3 & 83.9 & 50.0 & 64.3 \\
Skywork-Reward-Llama-3.1-8B  & 69.5 & 60.6 & 54.5 & 95.7 & 89.0 & 74.7 & 46.6 & 70.1 & 59.1 & 64.3 & 76.8 & 50.0 & 62.3 \\
Ray2333/GRM-llama3-8B-distill
                            & 62.4 & 62.1 & 56.9 & 88.1 & 82.2 & 71.5 & 48.4 & 67.4 & 57.1 & 66.3 & 78.6 & 54.8 & 62.0 \\
URM-Llama-3.1-8B             & 71.2 & 61.8 & 54.1 & 93.1 & 84.0 & 73.2 & 53.0 & 70.0 & 62.3 & 67.4 & 76.8 & 47.6 & 64.3 \\
\midrule

\rowcolor{qgray}
\multicolumn{14}{l}{\textbf{Phi-3.5-mini-instruct}} \\
\quad last token pooling & 54.9 & 53.1 & 53.4 & 70.8 & 79.1 & 60.2 & \textbf{34.7} & 58.0 
                         & 53.9 & 55.1 & 60.7 & 52.4 & 55.1 \\
\quad mean pooling      & 55.0 & 48.7 & 50.7 & 64.4 & 79.4 & 55.3 & 29.4 & 54.7 
                         & \textbf{55.8} & 55.1 & \textbf{62.5} & 47.6 & 55.7 \\
\rowcolor{qblue}
\quad \textbf{+ AdaJudge (ours)}     & \textbf{56.8} & \textbf{53.6} & \textbf{53.8} & \textbf{74.9} & \textbf{92.2} & \textbf{66.0} & 21.1 & \textbf{59.8}
                         & 54.6 & \textbf{66.3} & \textbf{62.5} & \textbf{57.1} & \textbf{59.4} \\
\midrule
\rowcolor{qgray}
\multicolumn{14}{l}{\textbf{Qwen3-4B}} \\
\quad + last token pooling & 61.5 & 60.6 & 67.6 & 80.8 & \textbf{93.6} & 74.3 & 35.0 & 67.6 & 57.1 & 62.2 & 76.8 & 61.9 & 62.3 \\
\quad + mean pooling      & 62.8 & 61.0 & \textbf{68.0} & \textbf{85.8} & 91.5 & 75.3 & 41.4 & 69.4 & 56.5 & \textbf{65.3} & 69.6 & \textbf{81.0} & 64.0 \\
\rowcolor{qblue}
\quad \textbf{+ AdaJudge (ours)}     & \textbf{68.3} & \textbf{68.9} & 61.6 & 84.5 & 91.9 & \textbf{76.7} & \textbf{43.7} & \textbf{70.8}
                          & \textbf{61.7} & 58.2 & \textbf{80.4} & \textbf{81.0} & \textbf{66.0} \\
\midrule

\rowcolor{qgray}
\multicolumn{14}{l}{\textbf{Qwen3-8B}} \\
\quad + last token pooling & 63.0 & 66.6 & 59.5 & 79.9 & 93.1 & 73.9 & 34.8 & 67.3 & \textbf{61.7} & 54.1 & 78.6 & 69.0 & 63.1 \\
\quad + mean pooling      & 60.6 & 66.0 & 60.0 & 82.6 & 86.0 & 72.2 & \textbf{43.6} & 67.3 & 55.2 & \textbf{64.3} & 76.8 & 73.8 & 63.4 \\
\rowcolor{qblue}
\quad \textbf{+ AdaJudge (ours)}     & \textbf{66.9} & \textbf{68.0} & \textbf{61.9} & \textbf{87.5} & \textbf{93.5} & \textbf{76.8} & 43.0 & \textbf{71.1}
                          & 58.4 & \textbf{64.3} & \textbf{80.4} & \textbf{78.6} & \textbf{66.0} \\

\bottomrule
\bottomrule
\end{tabular}
\end{adjustbox}

\caption[Performance of Reward Models on RM-Bench and JudgeBench]{
Performance of reward models on RM-Bench and JudgeBench (higher is better).
We report both strong off-the-shelf reward models trained on large-scale preference data and controlled baselines that share the same backbone and training setup but differ only in the reward modeling strategy.
This comparison isolates the impact of reward model architecture and aggregation design across model scales and domains.
Skywork-Reward-Llama-3.1-8B and Skywork-Reward-Gemma-2-27B~\citep{liu2024skywork} are the top models on the original RM-Bench~\citep{liu2024rm} and JudgeBench~\citep{judgebench2024} leaderboards respectively.}
\vspace{-8pt}
\label{tab:combined_rm_evaluation}
\end{table*}

%% file: chapters/experiment.tex
\section{Experiments}
\label{sec:exp}

In this section, we evaluate  AdaJudge by answering the following research questions (RQs):

\vspace{-5pt} \begin{itemize}[leftmargin=*]\setlength\itemsep{-0.3em} \item RQ1: How does the performance of AdaJudge compare to strong off-the-shelf reward models and traditional pooling baselines? (Section~\ref{sec:main_results})
\item RQ2: What are the trade-offs of different aggregation strategies, and how do static and adaptive readouts behave on complex, reasoning-intensive tasks? (Section~\ref{sec:ablation-study}) 

\item RQ3: How do the internal mechanisms
contribute to task-dependent preference discrimination? (Section~\ref{sec:analysis})
\end{itemize}

\subsection{Experimental Setup}
\label{sec:exp_setup}

\paragraph{Training Data and Models.}
All reward models are trained on the preference split of HelpSteer3, a human-annotated preference dataset constructed from real-world conversational prompts and annotated by qualified human raters across diverse tasks and languages, with about 40.5K preference pairs~\citep{wang2025helpsteer3preferenceopenhumanannotatedpreference}.
We experiment with 3
open-source base models: Phi-3.5-mini-instruct~\citep{abdin2024phi3}, Qwen3-4B and Qwen3-8B~\citep{qwen3}, and trained with LoRA. The Stage-I Iterative Refinement module consists of K small transformer blocks (K=2 for Phi-3.5-mini-instruct; K=3 for Qwen3-4B and Qwen3-8B) with a 2 feed-forward network layer. Additional training details are provided in Appendix~\ref{training_details}.

\paragraph{Baselines.}
Our evaluation considers two complementary perspectives. First, we report several widely used off-the-shelf reward models as reference points, and more details are given in Appendix~\ref{sec:appendix-more-details-baselines}. These classic models are trained on substantially larger and more diverse preference datasets than the single-dataset setting considered here, providing a strong indication of the performance level achieved by large-scale reward modeling approaches.
Second, to attribute performance differences specifically to architectural design, we construct controlled baselines under the same backbone, training data, and optimization function.
In this setting, AdaJudge's adaptive components are replaced with a fixed single-view readout, while keeping all other factors unchanged. All such baselines attach the same MLP reward head to map a pooled hidden representation to a scalar score: (i) last-hidden state, which uses the hidden state of the last non-padding token as the sequence representation before the MLP head; and (ii) mean pooling, which computes a masked mean of token hidden states over all non-padding tokens and scores it with the same MLP head.

\paragraph{Evaluation.}
To better evaluate reward models under recent and comprehensive judging benchmarks, we use RM-Bench~\citep{liu2024rm} and JudgeBench~\citep{judgebench2024}. RM-Bench measures pairwise judging accuracy across multiple domains (chat, math, code, and safety) and difficulty levels, with an emphasis on reducing style-driven artifacts~\citep{liu2024rm}.
JudgeBench provides complementary coverage with challenging preference pairs spanning knowledge, reasoning, math, and code, targeting judge reliability on more objective distinctions~\citep{judgebench2024}.
For both benchmarks, evaluation follows the official protocols and metrics provided by each benchmark.

\subsection{Main Results Analysis (RQ1)}
\label{sec:main_results}

Table~\ref{tab:combined_rm_evaluation} presents the comprehensive evaluation results on RM-Bench and JudgeBench. Our analysis yields three key observations.

\paragraph{Scaling Efficiency and Superior Performance.}
AdaJudge significantly amplifies the discriminative capacity of base models, enabling compact architectures to rival or surpass much larger, heavily optimized baselines. Notably, while the Qwen3-8B backbone with traditional fixed pooling falls short of the 27B-parameter Skywork-Reward-Gemma-2 (stagnating at $67.3$ vs. $70.5$ on RM-Bench), equipping it with AdaJudge pushes its overall performance to $71.1$, successfully surpassing the 27B model. This demonstrates that our method's performance dominance is not merely a byproduct of a strong base model. By explicitly decoupling representation refinement from evidence aggregation, AdaJudge breaks the artificial bottleneck of standard readouts, yielding greater parameter efficiency than brute-force scaling or massive data curation.

\paragraph{Alleviating Representation Bottlenecks in Compact Models.}
In parameter-constrained regimes like Phi-3.5-mini-instruct, standard fixed pooling forces a severe representational trade-off. Empirical results demonstrate a strict bifurcation: mean pooling degrades on localized reasoning tasks like Math and Code (scoring $48.7$ and $50.7$ on RM-Bench), while last-token pooling falters on holistic JudgeBench metrics ($55.1$ overall). Compact models inherently lack the latent dimensionality and attention capacity to flawlessly compress complex, diverse preference semantics into a single spatial view, leading to gradient conflicts during fine-tuning. AdaJudge directly relieves this representational burden. By offloading evidence extraction to explicit multi-perspective views via iterative refinement, it allows the limited backbone to maintain robust, general-purpose representations rather than over-optimizing for a single inductive bias. Consequently, this structural regularization yields a balanced and substantial lift across both benchmarks ($59.8$ on RM-Bench, $59.4$ on JudgeBench), proving that dynamic readouts can effectively compensate for intrinsic capacity limits without catastrophic forgetting.

\paragraph{Dynamic Resolution of Conflicting Inductive Biases.}
Granular domain analysis confirms the core structural limitation of static architectures: they impose a fixed receptive field that cannot generalize across heterogeneous tasks. For instance, in Qwen3-8B, mean pooling dominates holistic Safety evaluation ($82.6$ vs. $79.9$ for last-token), yet last-token pooling traditionally favors the strict, localized logic required for mathematical verification. Furthermore, standard pooling often collapses on highly ambiguous queries; for example, last-token pooling scores a mere $34.8$ on the RM-Bench Hard subset, highlighting its inability to capture subtle, distributed errors. AdaJudge overcomes this inherent tension via its prompt-aware routing mechanism. By conditioning the aggregation strategy on the input context, AdaJudge effectively translates a static architectural choice into a dynamic task-inference process. It retains high precision on Math ($80.4$), maximizes Safety ($87.5$), and significantly improves robustness on the Hard split ($43.0$). This achieves a true Pareto improvement, validating its capability to intelligently adapt its receptive field to the actual spatial distribution of verification evidence.

\input{tables/ablation1}
\input{tables/ablation2}


\subsection{Ablation Study (RQ2)}\label{sec:ablation-study}

\paragraph{Effectiveness of Adaptive Aggregation.} 
To isolate the effect of multi-perspective aggregation, we compare AdaJudge's dynamic approach against three static single-view baselines on Qwen3-4B: Last-Token, Mean-Pooling, and Attention-Pooling. In this setup, we fix the Stage-I refined representations and vary only the readout mechanisms.

As shown in Table~\ref{tab:ablation_final}, static pooling strategies exhibit conflicting performance profiles. On RM-Bench, Last-Token yields the strongest Safety score ($84.7\%$) but underperforms on Math ($65.0\%$). Conversely, Mean-Pooling improves Math ($68.0\%$) but incurs a substantial cost in Safety ($77.0\%$). 
This inversion on Math suggests that the Stage-I refinement effectively propagates localized reasoning signals across the sequence, creating a coupling effect that enables Mean-Pooling to leverage global context for cues typically confined to sparse tokens.
While Attention-Pooling offers a balanced compromise, it still lags behind AdaJudge. In contrast, AdaJudge employs an instance-adaptive mixture of views, achieving the best overall results on both RM-Bench ($70.8\%$) and JudgeBench ($66.0\%$). Notably, the gains are concentrated on complex instances: on the RM-Bench Hard subset, AdaJudge scores $43.7\%$, outperforming the best static alternative (Attention-Pooling, $38.7\%$) by a significant margin of $5.0\%$. This pattern demonstrates that fixed readouts impose a structural bottleneck, whereas adaptive aggregation accommodates heterogeneous evaluation criteria without committing to a single rigid inductive bias.

\paragraph{Representation Refinement for Discriminative Scoring.}
We further examine the role of Stage-I refinement using Qwen3-4B as the backbone, by comparing AdaJudge with a variant that bypasses this module (w/o Refinement), directly using backbone hidden states for scoring. As detailed in Table~\ref{tab:ablation_refinement}, incorporating refinement improves overall JudgeBench accuracy by $3.4\%$ ($62.6\% \to 66.0\%$). The gains are most pronounced in reasoning-intensive subsets, such as Math ($+5.4\%$) and Code ($+4.8\%$), where judgments rely on fine-grained logical consistency rather than surface-level patterns. On RM-Bench, both variants perform comparably on the Easy subset ($\approx 92\%$), suggesting raw representations suffice for straightforward comparisons. However, on the Hard subset, enabling refinement boosts accuracy from $39.6\%$ to $43.7\%$. These results confirm that while pretrained backbones encode coarse semantic signals, the adaptive refinement acts as a crucial "lens", reshaping representations into a discriminative space essential for complex reasoning.

%% file: tables/ablation1.tex
\begin{table*}[t]
\centering
\small
\setlength{\tabcolsep}{3.8pt}
\begin{tabular}{l|cccccccc|ccccc}
\toprule
& \multicolumn{8}{c|}{\textbf{RM-Bench}} & \multicolumn{5}{c}{\textbf{JudgeBench}} \\
\textbf{Method} & Chat & Math & Code & Safe & Easy & Norm & Hard & \textbf{Avg} & Know & Reas & Math & Code & \textbf{Avg} \\
\midrule
Last-Token & 66.0 & 65.0 & 61.2 & \textbf{84.7} & \textbf{94.6} & 76.1 & 37.0 & 69.2 & 57.1 & 66.3 & 75.0 & 78.6 & 65.1 \\
Mean-Pool & 64.7 & 68.0 & 59.9 & 77.0 & 92.5 & 74.9 & 34.8 & 67.4 & 58.4 & 62.2 & 78.6 & 71.4 & 64.3 \\
Attn-Pool & 66.6 & 68.0 & \textbf{62.1} & 81.4 & 93.0 & \textbf{76.9} & 38.7 & 69.5 & \textbf{61.7} & 57.1 & 76.8 & \textbf{81.0} & 65.1 \\
\midrule
\textbf{AdaJudge (Ours)} & \textbf{68.3} & \textbf{68.9} & 61.6 & 84.5 & 91.9 & 76.8 & \textbf{43.7} & \textbf{70.8} & \textbf{61.7} & \textbf{58.2} & \textbf{80.4} & \textbf{81.0} & \textbf{66.0} \\
\bottomrule
\end{tabular}
\caption{Ablation Study on Aggregation Strategies. We compare AdaJudge with static single-view baselines across all metrics on Qwen3-4B. Note that AdaJudge achieves the best overall performance on both benchmarks, especially on the Hard subset of RM-Bench.}
\label{tab:ablation_final}
\end{table*}

%% file: tables/ablation2.tex
\begin{table*}[t]
\centering
\small
\setlength{\tabcolsep}{3.8pt} 
\begin{tabular}{l|cccccccc|ccccc}
\toprule
& \multicolumn{8}{c|}{\textbf{RM-Bench}} & \multicolumn{5}{c}{\textbf{JudgeBench}} \\
\textbf{Method} & Chat & Math & Code & Safe & Easy & Norm & Hard & \textbf{Avg} & Know & Reas & Math & Code & \textbf{Avg} \\
\midrule
w/o Refinement & 63.7 & 68.7 & \textbf{61.8} & \textbf{84.7} & \textbf{92.5} & \textbf{77.2} & 39.6 & 69.8 & 55.8 & \textbf{60.2} & 75.0 & 76.2 & 62.6 \\
\midrule
\textbf{AdaJudge (Ours)} & \textbf{68.3} & \textbf{68.9} & 61.6 & 84.5 & 91.9 & 76.8 & \textbf{43.7} & \textbf{70.8} & \textbf{61.7} & 58.2 & \textbf{80.4} & \textbf{81.0} & \textbf{66.0} \\
\bottomrule
\end{tabular}
\caption{Impact of Adaptive Representation Refinement. Comparison between directly using Qwen3-4B backbone features (w/o Refinement) and applying our Stage-I refinement module. The refinement stage yields significant gains on complex tasks (e.g., Hard subset, Math, and Code).}
\label{tab:ablation_refinement}
\end{table*}

%% file: chapters/analysis.tex
\subsection{Internal Mechanism Analysis (RQ3)}
\label{sec:analysis}

In this section, we probe the internal mechanisms of AdaJudge to validate our motivating hypotheses regarding task-dependent inductive biases and representational refinement.
All analyses are conducted on RM-Bench and JudgeBench using the Qwen3-4B–based AdaJudge model.

\begin{figure}[t]
    \centering
     \includegraphics[width=\linewidth]{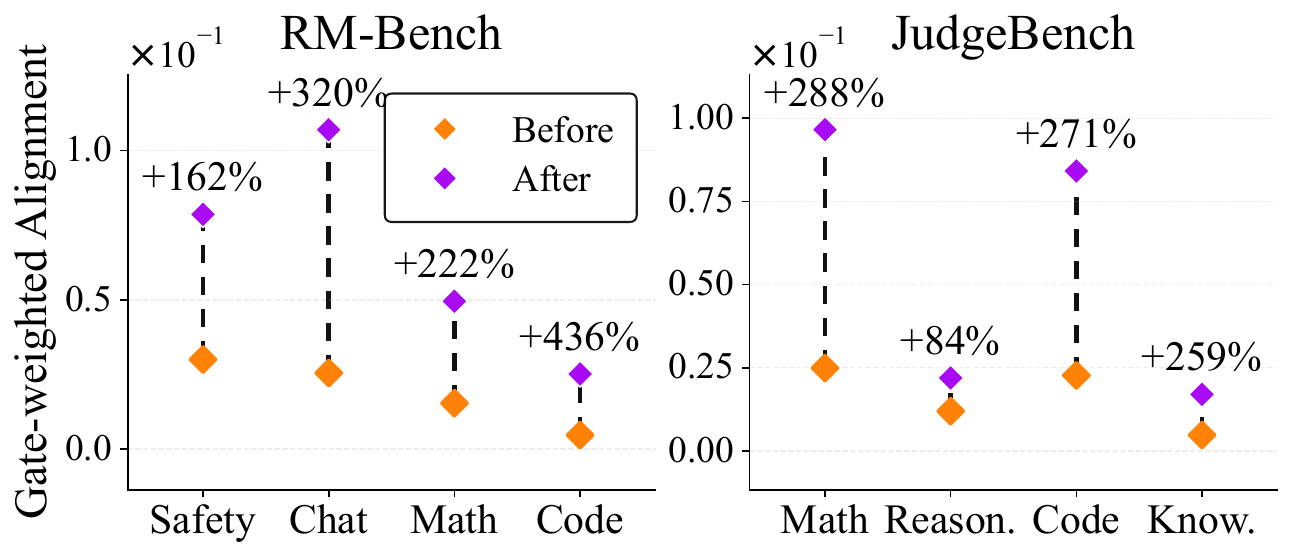}
    \caption{Gate-weighted alignment before and after Stage-I refinement on RM-Bench (left) and JudgeBench (right).
Alignment is measured as the cosine similarity between the scoring direction $\nabla_{\mathbf{z}} r(\mathbf{x},\mathbf{y})$ and the preference difference $\mathbf{z}(x^+)-\mathbf{z}(x^-)$, aggregated using AdaJudge’s routing weights.
}
\vspace{-8pt}
    \label{fig:analysis1}
\end{figure}

\paragraph{Does iterative refinement improve the representation space?}
We investigate whether iterative refinement improves the internal representation space in a way that is directly beneficial for preference discrimination.
For each preference pair $(x^+, x^-)$, we measure the cosine alignment between the preference difference and the reward model’s scoring direction to quantify the consistency between the latent preference shift and the optimization direction,
$\cos\!\big(\nabla_{\mathbf{z}} r(\mathbf{x},\mathbf{y}),\, \mathbf{z}(x^+) - \mathbf{z}(x^-)\big)$,
where $\mathbf{z}$ denotes a pooled representation produced by one of the three aggregation views defined in Section~\ref{sec:stage2}. 
We compute this alignment using pooled representations derived from the backbone features $\mathbf{H}^{(0)}$ (before refinement) and from the refined features $\tilde{\mathbf{H}}$ (after refinement), and additionally report a gate-weighted overall alignment that aggregates the three pooling views using AdaJudge’s routing weights.
As shown in Figure~\ref{fig:analysis1}, iterative refinement consistently improves alignment across domains, with the largest gains on discrimination-intensive settings such as Chat and Safety on RM-Bench, and Math and Code on JudgeBench.
Notably, alignment gains concentrate on mean and attention pooling views, while last-token alignment remains flat or even degrades, indicating that refinement makes preference-relevant evidence more globally accessible rather than sharpening a single terminal representation.
These results show that iterative refinement functions as a discrimination-oriented lens, reshaping representations so that AdaJudge’s scoring direction better matches true preference-separating directions, particularly under global and sparse-evidence aggregation.

\paragraph{Do distinct domains elicit specific aggregation patterns?}

\begin{figure}[t]
    \centering
     \includegraphics[width=\linewidth]{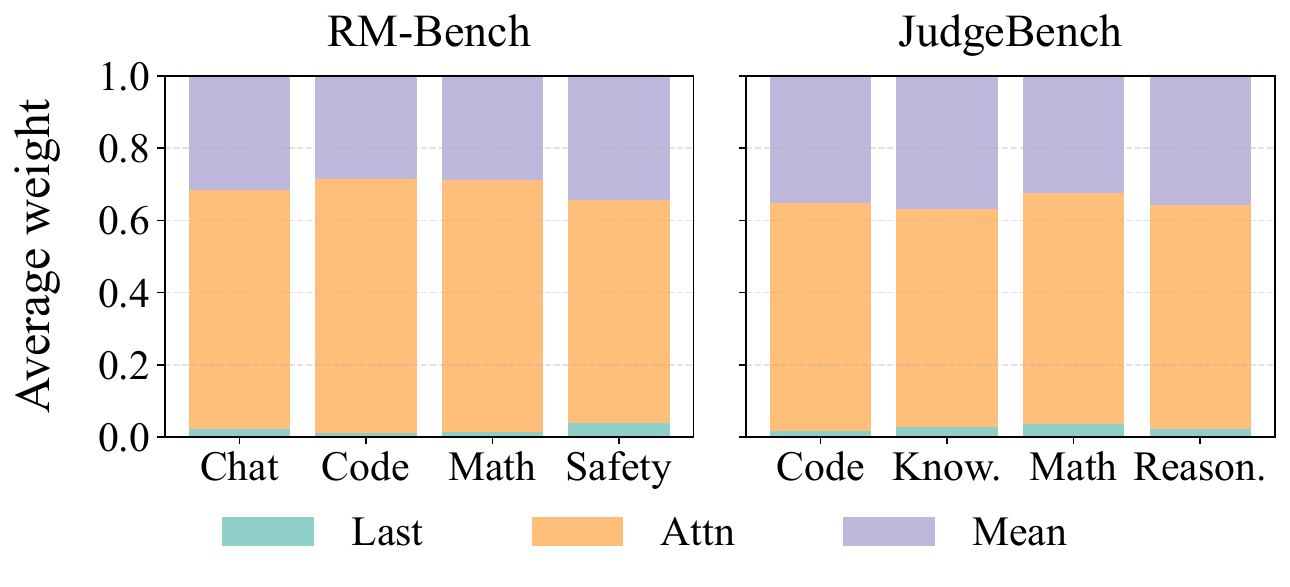}
    \caption{Average routing weights of AdaJudge’s pooling experts across domains on RM-Bench (left) and JudgeBench (right).
For each benchmark, we group pairwise preference samples by domain and compute the mean gating weights assigned to the Last, Attention, and Mean pooling experts over the responses, using a Qwen3-4B–based AdaJudge reward model.
}
    \label{fig:analysis2}
\vspace{-8pt}
\end{figure}

We analyze AdaJudge’s aggregation behavior by examining the routing distributions $\boldsymbol{\pi}$ defined in Section~\ref{sec:stage2}.
For each benchmark, we group preference pairs by domain and compute the average routing weights assigned to the Last, Attention, and Mean pooling experts over the chosen responses.
The resulting distributions are shown in Figure~\ref{fig:analysis2}, which indicate that distinct and structured aggregation patterns emerge across domains.
On Code and Math, AdaJudge consistently allocates a larger proportion of weight to the Attention pooling expert, whereas the Safety domain exhibits a clear shift toward increased reliance on Mean pooling. In contrast, the Last-token expert receives a relatively small average weight across all domains.
This behavior suggests that terminal-token representations alone are generally insufficient for preference discrimination once richer token-level evidence is made available, and that much of the functionality traditionally attributed to last-token pooling is subsumed by attention-based aggregation. Overall, these results indicate that AdaJudge does not apply a uniform aggregation strategy across tasks. Instead, distinct domains elicit systematically different routing behaviors, demonstrating that the model adapts its evidence aggregation to domain-dependent evaluation characteristics under a single unified architecture.

%% file: chapters/conclusion.tex
\section{Conclusions and Future Work}
This work examines a core architectural assumption in reward modeling and shows relying on fixed aggregation over generative representations limits effective preference discrimination. To overcome this rigidity, we introduce AdaJudge, a unified framework that jointly adapts the representation space and the aggregation strategy. 
Experiments on RM-Bench and JudgeBench confirm that this two-stage adaptation yields clear improvements over strong reward models and fixed-pooling baselines, including in same-backbone comparisons. 
In future we plan to explore extending the AdaJudge framework to multi-objective reward modeling, enabling the routing mechanism to better disentangle complex trade-offs between diverse criteria, such as factuality and helpfulness.

%% file: chapters/appendix.tex
\appendix

\section{Training Details.}
\label{training_details}
We train all reward models using a preference-based objective with a focal Bradley--Terry loss~\citep{bradley1952rank}.
Training maximum sequence length is 4096 tokens and gradient checkpointing enabled to reduce memory consumption.
We apply LoRA with rank $r{=}96$, scaling factor $\alpha{=}128$, and dropout $0.05$. Gradient norms are clipped to a maximum value of $2.0$.
We use the AdamW optimizer with $\beta_1{=}0.9$, $\beta_2{=}0.95$, and weight decay $0.1$. A constant learning rate schedule with linear warmup is adopted, where the warmup ratio is set to $3\%$ of the total training steps. Checkpoints are saved every 25 optimization steps.

For Qwen3-4B, we use a learning rate of $4{\times}e^{-5}$ and set the number of iterative refinement blocks to $\mathrm{K}{=}3$.
The focal Bradley--Terry loss uses temperature $\mathrm{T}{=}1.3$ and focusing parameter $\gamma{=}0.9$.
The pooling head employs an entropy regularization coefficient of $0.01$ with a target gate entropy of $0.7$.

For Phi-3.5-mini-instruct, we adopt a learning rate of $2{\times}e^{-5}$ and use $\mathrm{K}{=}2$ iterative refinement blocks.
The focal loss temperature and focusing parameters are set to $\mathrm{T}{=}1.2$ and $\gamma{=}0.5$, respectively.
The pooling head uses the same entropy regularization coefficient of $0.01$ and a target entropy of $0.7$.

For Qwen3-8B, we reduce the learning rate to $3{\times}e^{-5}$ and set $\mathrm{K}{=}3$ refinement blocks.
The focal Bradley--Terry loss uses $\mathrm{T}{=}1.2$ and $\gamma{=}0.7$.
We apply a smaller entropy regularization coefficient of $0.003$ with a target gate entropy of $0.65$ for the pooling head.

All models are trained with an effective batch size of 8 preference pairs on a single NVIDIA RTX Pro 6000 GPU.
Table \ref{tab:compute_cost} details the training overhead. Compared to standard pooling baselines, AdaJudge incurs additional memory footprint and training time due to the iterative refinement and dynamic routing mechanisms. However, the peak memory usage ($32.99$ GB for Qwen3-8B) remains well within the capacity of modern GPUs, and the maximum training time of $10.80$ hours is highly manageable. The slightly lower throughput (Train TFLOP/h) reflects the sequential nature of our routing mechanism, which underutilizes raw GPU compute compared to dense forward passes.

\definecolor{qgray}{RGB}{240,240,240}
\definecolor{qblue}{RGB}{230,245,255}

\begin{table}[htbp]
\centering
\small 
\renewcommand{\arraystretch}{1.1} 
\setlength{\tabcolsep}{3pt} 
\begin{tabular}{l|rrr}
\toprule
\textbf{Method} & \textbf{TFLOP/h} & \textbf{Mem (GB)} & \textbf{Time (h)} \\
\midrule

\rowcolor{qgray}
\multicolumn{4}{l}{\textbf{Phi-3.5-mini-instruct}} \\
\quad last token & $6.05 \times 10^5$ & 13.15 & 3.56 \\
\quad mean pooling & $6.14 \times 10^5$ & 13.15 & 3.51 \\
\quad only stage 1 & $4.12 \times 10^5$ & 18.12 & 7.21 \\
\quad only stage 2 & $4.14 \times 10^5$ & 13.64 & 6.77 \\
\rowcolor{qblue}
\quad \textbf{AdaJudge (ours)} & $3.57 \times 10^5$ & 16.80 & 7.71 \\
\midrule

\rowcolor{qgray}
\multicolumn{4}{l}{\textbf{Qwen3-4B}} \\
\quad last token & $3.95 \times 10^5$ & 14.85 & 5.39 \\
\quad mean pooling & $3.98 \times 10^5$ & 14.85 & 5.33 \\
\quad only stage 1 & $3.72 \times 10^5$ & 17.63 & 7.20 \\
\quad only stage 2 & $3.72 \times 10^5$ & 15.19 & 6.92 \\
\rowcolor{qblue}
\quad \textbf{AdaJudge (ours)} & $3.12 \times 10^5$ & 18.61 & 8.08 \\
\midrule

\rowcolor{qgray}
\multicolumn{4}{l}{\textbf{Qwen3-8B}} \\
\quad last token & $5.30 \times 10^5$ & 24.52 & 7.51 \\
\quad mean pooling & $5.30 \times 10^5$ & 24.52 & 7.51 \\
\quad only stage 1 & $5.14 \times 10^5$ & 31.02 & 9.93 \\
\quad only stage 2 & $5.16 \times 10^5$ & 25.40 & 9.40 \\
\rowcolor{qblue}
\quad \textbf{AdaJudge (ours)} & $4.46 \times 10^5$ & 32.99 & 10.80 \\

\bottomrule
\end{tabular}
\vspace{-4pt}
\caption{Training overhead comparison on a single NVIDIA RTX Pro 6000 GPU.}
\label{tab:compute_cost}
\end{table}

\section{More Details of the External Baselines}
\label{sec:appendix-more-details-baselines}

\paragraph{Skywork-Reward-Gemma-2-27B.}
A 27B reward model built on gemma-2-27b-it, trained on 80K curated public preference pairs(Skywork Reward Data Collection) to predict human preference scores in a Bradley–Terry setup. It leads the RewardBench leaderboard with strong multi-domain performance~\citep{liu2024skywork}.

\paragraph{Skywork-Reward-Llama-3.1-8B.}
An 8B reward model based on the Meta-Llama-3.1-8B-Instruct backbone. It is trained under the same data and optimization objective setup as Skywork-Reward-Gemma-2-27B, differing primarily in backbone scale. Despite its smaller size, it serves as a strong public reward-model baseline~\citep{liu2024skywork}.

\paragraph{Ray2333/GRM-llama3-8B-distill.}
An 8B reward model from the GRM collection that fine-tunes only a linear reward head on Llama3-8B pretrained weights using the preference 700K dataset~\citep{dong2024rlhf}. This simple but competitive strategy yields a lightweight scoring RM~\citep{yang2024regularizing}.
\paragraph{URM-LLaMa-3.1-8B.}
An uncertainty-aware reward model that extends a Skywork-Reward-Llama-3.1-8B base with an uncertainty-aware and attribute-specific value head, learning to combine multiple fine-grained quality attributes into a final reward signal\citep{lou2025uncertaintyawarerewardmodelteaching}.

\section{More Details of Benchmark Datasets}\label{sec:appendix-more-details-benchmarks}
\paragraph{RM-Bench.}
RM-Bench is a benchmark designed to evaluate reward models under subtle content differences and stylistic variations~\citep{liu2024rm}.
This benchmark contains 1.33k test instances. Each test instance consists of a single prompt paired with three chosen responses and three rejected responses, corresponding to three distinct styles: concise, detailed plain text, and detailed markdown.
The dataset covers multiple domains, including chat, code, math, safety-refuse, and safety-response.

Evaluation on RM-Bench is performed by comparing the scores of chosen and rejected responses in a $3\times3$ manner for each instance, resulting in nine pairwise comparisons per prompt.
Based on the relative styles of the compared responses, three accuracy metrics are reported:
Hard accuracy evaluates cases where a chosen response has a simpler style than the rejected one,
Normal accuracy compares responses with the same style,
and Easy accuracy corresponds to cases where the chosen response has a more elaborate style.
This evaluation protocol explicitly tests a reward model's robustness to stylistic bias rather than surface-level preference.

\paragraph{JudgeBench.}
JudgeBench is a benchmark for evaluating LLM-based judges on objective correctness rather than subjective preference~\citep{judgebench2024}.
Each instance consists of a question paired with two candidate responses, where one response is objectively correct and the other is incorrect, as determined by ground-truth labels.

The benchmark contains two official test splits: a gpt split with 350 response pairs generated by GPT-4o~\citep{openai2024gpt4o}, and a claude split with 270 response pairs generated by Claude-3.5-Sonnet~\citep{anthropic2024claude35sonnet}.
The evaluation spans multiple domains, including knowledge, reasoning, math, and coding.
Performance is measured by the accuracy with which a judge model ranks the correct response above the incorrect one.

%% file: custom.bib
@article{lin2017focal,
  title   = {Focal Loss for Dense Object Detection},
  author  = {Lin, Tsung-Yi and Goyal, Priya and Girshick, Ross and He, Kaiming and Doll{\'a}r, Piotr},
  journal = {arXiv preprint arXiv:1708.02002},
  year    = {2017}
}

@article{liu2024rm,
  title={RM-Bench: Benchmarking Reward Models of Language Models with Subtlety and Style},
  author={Liu, Yantao and Yao, Zijun and Min, Rui and Cao, Yixin and Hou, Lei and Li, Juanzi},
  journal={arXiv preprint arXiv:2410.16184},
  year={2024}
}

@misc{judgebench2024,
  title={JudgeBench: A Benchmark for Evaluating LLM-Based Judges},
  author={Sijun Tan and Siyuan Zhuang and Kyle Montgomery and Willian Yuan Tang and Alejandro Cuadron and Chenguang Wang and Raluca Ada Popa and Ion Stoica},
  year={2024},
  archivePrefix={arXiv},
  url={https://arxiv.org/abs/2410.12784}
}

@article{qwen3,
  title = {Qwen3 Technical Report},
  author = {Yang, An and Li, Anfeng and Yang, Baosong and Zhang, Beichen and Hui, Binyuan and others},
  year = {2025},
  journal = {arXiv preprint arXiv:2505.09388}
}

@inproceedings{christiano2017deep,
  title={Deep reinforcement learning from human preferences},
  author={Christiano, Paul F and Leike, Jan and Brown, Tom and Martic, Miljan and Legg, Shane and Amodei, Dario},
  booktitle={Advances in Neural Information Processing Systems},
  volume={30},
  year={2017}
}

@inproceedings{ouyang2022training,
  title={Training language models to follow instructions with human feedback},
  author={Ouyang, Long and Wu, Jeffrey and Jiang, Xu and Almeida, Diogo and Wainwright, Carroll L and others},
  booktitle={Advances in Neural Information Processing Systems},
  volume={35},
  pages={27730--27744},
  year={2022}
}

@article{grattafiori2024llama3herdmodels,
    title   = {The Llama 3 Herd of Models},
  author  = {Grattafiori, Aaron and Dubey, Abhimanyu and Jauhri, Abhinav and Pandey, Abhinav and Kadian, Abhishek and others},
  journal = {arXiv preprint arXiv:2407.21783},
  year    = {2024}
}

@article{achiam2023gpt,
  title={GPT-4 Technical Report},
  author={Achiam, Josh and Adler, Steven and Agarwal, Sandhini and Ahmad, Lama and Akkaya, Ilge and others},
  journal={arXiv preprint arXiv:2303.08774},
  year={2023}
}

@misc{wang2025helpsteer3preferenceopenhumanannotatedpreference,
      title={HelpSteer3-Preference: Open Human-Annotated Preference Data across Diverse Tasks and Languages}, 
      author={Zhilin Wang and Jiaqi Zeng and Olivier Delalleau and Hoo-Chang Shin and Felipe Soares and Alexander Bukharin and Ellie Evans and Yi Dong and Oleksii Kuchaiev},
      year={2025},
      eprint={2505.11475},
      archivePrefix={arXiv},
      primaryClass={cs.CL},
      url={https://arxiv.org/abs/2505.11475}, 
}

@article{cui2023ultrafeedback,
      title   = {UltraFeedback: Boosting Language Models with Scaled AI Feedback},
  author  = {Cui, Ganqu and Yuan, Lifan and Ding, Ning and Yao, Guanming and He, Bingxiang and Zhu, Wei and Ni, Yuan and Xie, Guotong and Xie, Ruobing and Lin, Yankai and Liu, Zhiyuan and Sun, Maosong},
  journal = {arXiv preprint arXiv:2310.01377},
  year    = {2023}
}

@misc{lightman2023let,
      title={Let's Verify Step by Step}, 
      author={Hunter Lightman and Vineet Kosaraju and Yura Burda and Harri Edwards and Bowen Baker and Teddy Lee and Jan Leike and John Schulman and Ilya Sutskever and Karl Cobbe},
      year={2023},
      eprint={2305.20050},
      archivePrefix={arXiv},
      primaryClass={cs.LG},
      url={https://arxiv.org/abs/2305.20050}, 
}

@article{luo2023wizardmath,
  title={WizardMath: Empowering Mathematical Reasoning for Large Language Models via Reinforced Evol-Instruct},
  author={Luo, Haipeng and Sun, Qingfeng and Xu, Can and Zhao, Pu and Lou, Jianguang and Tao, Chongyang and Geng, Xiubo and Lin, Qingwei and Chen, Shifeng and Zhang, Dongmei},
  journal={arXiv preprint arXiv:2308.09583},
  year={2023}
}

@inproceedings{reimers2019sentence,
  title     = {Sentence-BERT: Sentence Embeddings using Siamese BERT-Networks},
  author    = {Reimers, Nils and Gurevych, Iryna},
  booktitle = {Proceedings of the 2019 Conference on Empirical Methods in Natural Language Processing and the 9th International Joint Conference on Natural Language Processing (EMNLP-IJCNLP)},
  pages     = {3982--3992},
  year      = {2019},
  url       = {https://aclanthology.org/D19-1410/}
}

@article{liu2025skywork,
  title={Skywork-Reward-V2: Scaling Preference Data Curation via Human-AI Synergy},
  author = {Liu, Chris Yuhao and Zeng, Liang and Xiao, Yuzhen and He, Jujie and Liu, Jiacai and Wang, Chaojie and Yan, Rui and Shen, Wei and Zhang, Fuxiang and Xu, Jiacheng and Liu, Yang and Zhou, Yahui},
  journal={arXiv preprint arXiv:2507.01352},
  year={2025}
}

@article{wang2024armorm,
  title        = {Interpretable Preferences via Multi-Objective Reward Modeling and Mixture-of-Experts},
  author       = {Wang, Haoxiang and Xiong, Wei and Xie, Tengyang and Zhao, Han and Zhang, Tong},
  journal      = {arXiv preprint arXiv:2406.12845},
  year         = {2024},
  eprint       = {2406.12845},
  archivePrefix= {arXiv},
  primaryClass = {cs.LG}
}

@inproceedings{zhu2024starling,
  title     = {Starling-7B: Improving Helpfulness and Harmlessness with RLAIF},
  author    = {Zhu, Banghua and Frick, Evan and Wu, Tianhao and Zhu, Hanlin and Ganesan, Karthik and Chiang, Wei-Lin and Zhang, Jian and Jiao, Jiantao},
  booktitle = {Conference on Language Modeling (COLM)},
  year      = {2024},
  url       = {https://openreview.net/forum?id=GqDntYTTbk}
}

@article{cobbe2021training,
  title        = {Training Verifiers to Solve Math Word Problems},
  author       = {Cobbe, Karl and Kosaraju, Vineet and Bavarian, Mohammad and Chen, Mark and Jun, Heewoo and Kaiser, Lukasz and Plappert, Matthias and Tworek, Jerry and Hilton, Jacob and Nakano, Reiichiro and Hesse, Christopher and Schulman, John},
  journal      = {arXiv preprint arXiv:2110.14168},
  year         = {2021},
  eprint       = {2110.14168},
  archivePrefix= {arXiv}
}

@article{coste2023reward,
  title   = {Reward Model Ensembles Help Mitigate Overoptimization},
  author  = {Coste, Thomas and Anwar, Usman and Kirk, Robert and Krueger, David},
  journal = {arXiv preprint arXiv:2310.02743},
  year    = {2023}
}

@inproceedings{bengio2013deep,
  title     = {Deep Learning of Representations: Looking Forward},
  author    = {Bengio, Yoshua},
  booktitle = {Statistical Language and Speech Processing (SLSP 2013)},
  series    = {Lecture Notes in Computer Science},
  volume    = {7978},
  pages     = {1--37},
  year      = {2013},
  publisher = {Springer},
  doi       = {10.1007/978-3-642-39593-2_1}
}

@article{raposo2024mixture,
  title        = {Mixture-of-Depths: Dynamically allocating compute in transformer-based language models},
  author       = {Raposo, David and Ritter, Sam and Richards, Blake and Lillicrap, Timothy and Conway Humphreys, Peter and Santoro, Adam},
  journal      = {arXiv preprint arXiv:2404.02258},
  year         = {2024},
  eprint       = {2404.02258},
  archivePrefix= {arXiv}
}

@inproceedings{elbayad2020depth,
  title     = {Depth-Adaptive Transformer},
  author    = {Elbayad, Maha and Gu, Jiatao and Grave, Edouard and Auli, Michael},
  booktitle = {International Conference on Learning Representations (ICLR)},
  year      = {2020},
  url       = {https://openreview.net/forum?id=SJg7KhVKPH},
  eprint    = {1910.10073},
  archivePrefix= {arXiv},
  primaryClass = {cs.CL}
}

@article{ziegler2019finetuning,
title   = {Fine-Tuning Language Models from Human Preferences},
  author  = {Ziegler, Daniel M. and Stiennon, Nisan and Wu, Jeffrey and Brown, Tom B. and Radford, Alec and Amodei, Dario and Christiano, Paul and Irving, Geoffrey},
  journal = {arXiv preprint arXiv:1909.08593},
  year    = {2019}
}

@misc{stiennon2020summarize,
      title={Learning to summarize from human feedback}, 
      author={Nisan Stiennon and Long Ouyang and Jeff Wu and Daniel M. Ziegler and Ryan Lowe and Chelsea Voss and Alec Radford and Dario Amodei and Paul Christiano},
      year={2020},
      eprint={2009.01325},
      archivePrefix={arXiv},
      primaryClass={cs.CL},
}

@misc{bai2022constitutional,
 title={Constitutional AI: Harmlessness from AI Feedback}, 
      author={Yuntao Bai and Saurav Kadavath and Sandipan Kundu and Amanda Askell and Jackson Kernion and Andy Jones and Anna Chen and Anna Goldie and Azalia Mirhoseini and Cameron McKinnon and Carol Chen and Catherine Olsson and Christopher Olah and Danny Hernandez and Dawn Drain and Deep Ganguli and Dustin Li and Eli Tran-Johnson and Ethan Perez and Jamie Kerr and Jared Mueller and Jeffrey Ladish and Joshua Landau and Kamal Ndousse and Kamile Lukosuite and Liane Lovitt and Michael Sellitto and Nelson Elhage and Nicholas Schiefer and Noemi Mercado and Nova DasSarma and Robert Lasenby and Robin Larson and Sam Ringer and Scott Johnston and Shauna Kravec and Sheer El Showk and Stanislav Fort and Tamera Lanham and Timothy Telleen-Lawton and Tom Conerly and Tom Henighan and Tristan Hume and Samuel R. Bowman and Zac Hatfield-Dodds and Ben Mann and Dario Amodei and Nicholas Joseph and Sam McCandlish and Tom Brown and Jared Kaplan},
      year={2022},
      eprint={2212.08073},
      archivePrefix={arXiv},
      primaryClass={cs.CL},
}

@article{rafailov2023dpo,
  title   = {Direct Preference Optimization: Your Language Model is Secretly a Reward Model},
  author  = {Rafailov, Rafael and Sharma, Archit and Mitchell, Eric and Ermon, Stefano and Manning, Christopher D. and Finn, Chelsea},
  journal = {arXiv preprint arXiv:2305.18290},
  year    = {2023}
}

@article{abdin2024phi3,
  title         = {Phi-3 Technical Report: A Highly Capable Language Model Locally on Your Phone},
  author        = {Abdin, Marah and Aneja, Jyoti and Awadalla, Hany and Awadallah, Ahmed and others},
  journal       = {arXiv preprint arXiv:2404.14219},
  year          = {2024},
  eprint        = {2404.14219},
  archivePrefix = {arXiv},
  primaryClass  = {cs.CL}
}

@misc{liu2024skywork,
      title={Skywork-Reward: Bag of Tricks for Reward Modeling in LLMs}, 
      author={Chris Yuhao Liu and Liang Zeng and Jiacai Liu and Rui Yan and Jujie He and Chaojie Wang and Shuicheng Yan and Yang Liu and Yahui Zhou},
      year={2024},
      eprint={2410.18451},
      archivePrefix={arXiv},
      primaryClass={cs.AI},
}

@misc{yang2024regularizing,
      title={Regularizing Hidden States Enables Learning Generalizable Reward Model for LLMs}, 
      author={Rui Yang and Ruomeng Ding and Yong Lin and Huan Zhang and Tong Zhang},
      year={2024},
      eprint={2406.10216},
      archivePrefix={arXiv},
      primaryClass={cs.CL},
}

@misc{dong2024rlhf,
      title={RLHF Workflow: From Reward Modeling to Online RLHF}, 
      author={Hanze Dong and Wei Xiong and Bo Pang and Haoxiang Wang and Han Zhao and Yingbo Zhou and Nan Jiang and Doyen Sahoo and Caiming Xiong and Tong Zhang},
      year={2024},
      eprint={2405.07863},
      archivePrefix={arXiv},
      primaryClass={cs.LG},
}

@misc{lou2025uncertaintyawarerewardmodelteaching,
      title={Uncertainty-aware Reward Model: Teaching Reward Models to Know What is Unknown}, 
      author={Xingzhou Lou and Dong Yan and Wei Shen and Yuzi Yan and Jian Xie and Junge Zhang},
      year={2025},
      eprint={2410.00847},
      archivePrefix={arXiv},
      primaryClass={cs.LG},
}

@misc{anthropic2024claude35sonnet,
  title        = {Claude 3.5 Sonnet Model Card},
  author       = {{Anthropic}},
  year         = {2024},
  howpublished = {\url{https://www.anthropic.com/news/claude-3-5-sonnet}},
  note         = {Accessed: 2025-01}
}

@misc{openai2024gpt4o,
  title        = {GPT-4o System Card},
  author       = {{OpenAI}},
  year         = {2024},
  howpublished = {\url{https://openai.com/index/gpt-4o-system-card/}},
  note         = {Accessed: 2025-01}
}

@article{ennadir2025poolwisely,
  title={Pool Me Wisely: On the Effect of Pooling in Transformer-Based Models},
  author={Ennadir, Sofiane and others},
  journal={arXiv preprint arXiv:2510.03339},
  year={2025}
}

@inproceedings{wang2024interpretable,
  title={Interpretable Preferences via Multi-Objective Reward Modeling and Mixture-of-Experts},
  author={Wang, Haoxiang and Xiong, Wei and others},
  booktitle={Findings of the Association for Computational Linguistics (ACL)},
  year={2024}
}

@article{poolingattention2024,
  title={Pooling And Attention: What Are Effective Designs For LLM-Based Embedding Models?},
  author={Wang, J. and others},
  journal={OpenReview},
  year={2024}
}

@inproceedings{askstrongjudge2025,
  title={Ask a Strong LLM Judge when Your Reward Model is Uncertain},
  author={Xu, Zhenghao and Lu, Qin and Zhang, Qingru and Qiu, Liang and Hong, Ilgee and Yu, Changlong and Yao, Wenlin and Liu, Yao and Jiang, Haoming and Li, Lihong and others},
  booktitle={Advances in Neural Information Processing Systems (NeurIPS)},
  year={2025}
}

@inproceedings{lu2024routingexpert,
  title={Routing to the Expert: Efficient Reward-guided Ensemble of Large Language Models},
  author={Lu, Yuan and others},
  booktitle={Proceedings of the 2024 Conference of the North American Chapter of the Association for Computational Linguistics (NAACL)},
  year={2024}
}

@article{rewardmodelrouting2025,
  title={Reward Model Routing in Alignment},
  author={Wu, Xinle and Lu, Yao},
  journal={arXiv preprint arXiv:2510.02850},
  year={2025}
}

@article{brothers2025robust,
  title={Robust Noise Attenuation via Adaptive Pooling of Transformer Outputs},
  author={Brothers, John and others},
  journal={arXiv preprint arXiv:2506.09215},
  year={2025}
}

@article{bradley1952rank,
  title     = {Rank Analysis of Incomplete Block Designs: I. The Method of Paired Comparisons},
  author    = {Bradley, Ralph Allan and Terry, Milton E.},
  journal   = {Biometrika},
  volume    = {39},
  number    = {3/4},
  pages     = {324--345},
  year      = {1952}
}

@article{xiao2025prompt,
  title={Prompt-based adaptation in large-scale vision models: A survey},
  author={Xiao, Xi and Zhang, Yunbei and Zhao, Lin and Liu, Yiyang and Liao, Xiaoying and Mai, Zheda and Li, Xingjian and Wang, Xiao and Xu, Hao and Hamm, Jihun and others},
  journal={arXiv preprint arXiv:2510.13219},
  year={2025}
}

@inproceedings{lu2024routing,
  title={Routing to the Expert: Efficient Reward-Guided Ensemble of Large Language Models},
  author={Lu, Yuan and others},
  booktitle={Proceedings of NAACL},
  year={2024}
}
